\documentclass[10pt, a4paper]{article}
\usepackage{booktabs}
\usepackage{adjustbox}
\usepackage{listings}
\usepackage{xcolor}
\usepackage[final]{lrec2026}

\title{From Script to Semantics: Prompting Strategies for African NLI}

\name{Anuj Tiwari\textsuperscript{1}, Terry Oko-odion\textsuperscript{2}, Hannah Nwokocha\textsuperscript{3}} 

\address{
Noida Institute of Engineering and Technology\textsuperscript{1},
ML Collective\textsuperscript{1,2,3} \\
India\textsuperscript{1}, Nigeria\textsuperscript{2,3} \\
aj11anuj123@gmail.com, terryokoodion@gmail.com, hannahsopuruchi@gmail.com
}

\abstract{
Large language models (LLMs) are increasingly evaluated in multilingual settings, yet their inference behavior in low-resource African languages remains underexplored especially under pure prompting without fine-tuning. We present a systematic study of prompting strategies for Natural Language Inference (NLI) in Swahili, Yoruba, and Hausa using the AfriXNLI benchmark. We evaluate five prompting strategies Baseline (zero-shot), Script-Aware, Language Specific, Contrastive, and Native-Label Self-Translation (NL-STP) across two mid-sized open weight models (Llama3.2-3B and Gemma3-4B). To isolate the effect of prompt design, the effect of few-shot examples and Chain-of-Thought reasoning is eliminated in our study. We find a significant difference in performance of class wise across strategies with highly neutral class collapse and high prediction skew in some configurations. Contrastive prompting proves to be the most reliable and steadily improving strategy over language and model and has better balance of class behavior and balance of overall accuracy gains. Notably, well-constructed prompts are sufficient to beat more powerful baselines that are provided with few-shot prompts and Chain-of-Thought prompts. We have found that prompt formulation is essential to multilingual NLI with low-resource languages and that language aware decision structuring can be used to meaningfully enhance robustness in resource challenged settings.
 \\ \newline }

\begin{document}

\maketitleabstract

\section{Introduction}
Large language models (LLMs) have shown high efficiency in performing most natural language understanding tasks, however, their performance in low-resource multilingual conditions has been under-characterized. Specifically, Natural Language Inference (NLI) is one of the fundamental tasks that can be used to assess reasoning and semantic comprehension and has been most commonly investigated in high-resource languages. In most African languages, such as Swahili, Yoruba and Hausa, still only limited systematic studies of the behavior of LLM under regimes of pure prompting. Recent research in prompting has demonstrated that performance can be very sensitive to prompt formulation, instruction structure and reasoning scaffold. Nevertheless, the majority of research works concentrate on few-shot learning or Chain-of-Thought (CoT) prompting and do not pay much attention to the impact of carefully designed zero-shot prompts only on model behavior. This difference problem is particularly applicable in resource limited settings, where curated demonstrations, large scale fine-tuning or computationally prohibitive reasoning plans may not be practicable.

In this paper, we have a controlled study of prompting techniques of multilingual NLI in AfriXNLI benchmark \cite{afrixnli_dataset} of Swahili, Yoruba and Hausa. We apply five prompting methods including Baseline (zero-shot), Script Aware prompting, Language Specific prompting, Contrastive prompting, and Native-Label Self-Translation (NL-STP) Prompting to two mid-sized open weight LLMs (Llama3.2-3B and Gemma3-4B). We are not interested in the state of the art performance, but are interested in the determination of the effect of prompt structure on the behavior of classes, their robustness and their predictability in low resource environments. 

The research questions that guide our study are as follows:

\begin{itemize}
    \item \textbf{RQ1}: What is the effect of prompt design on class wise inference behavior in low-resource African languages in conditions?
    \item \textbf{RQ2}: Are language conscious and contrastive prompting strategies useful in reducing prediction skew and neutral class collapse in multilingual NLI?
    \item \textbf{RQ3}: Are better structured prompts superior to more powerful baselines, which are augmented with few-shot examples and Chain-of-Thought reasoning in the resource limited setting?
\end{itemize}

Through systematic evaluation and detailed class wise analysis, we demonstrate that prompt formulation significantly affects inference dynamics, often altering prediction distributions and stability across languages. Contrasting prompting is the strategy that has been assessed as the most consistent and powerful one as it enhances balance within the entailment, contradiction and neutral classes. Our results demonstrate the significance of language aware prompt design to make sound inferences in multilingual setting especially in low resource African settings.

\section{Related Work}

\subsection{African Language Benchmarks and Multilingual Evaluation}
Recent large-scale evaluation efforts have highlighted persistent performance gaps between African languages and high-resource languages in large language models. IrokoBench \cite{adelani2024irokobench} presents assessment suites including AfriXNLI, AfriMGSM and AfriMMLU across 17 African languages and has been found to have significant degradation compared to English with differences of up to 45 points between tasks and languages. Likewise, AfroBench \cite{ojo2025afrobench} compares 64 languages of Africa on 15 tasks and demonstrates that proprietary models are much more successful than open models on the tasks, and prompted LLMs tend to be less successful than supervised systems like AfroXLMR \cite{alabi2022afroxlmr} and AfriTeVa \cite{oladipo2023afriteva} in the situations where supervised data exists.

While these benchmarks provide performance comparison in a broad way, they a small number of prompt templates do not systematically design the prompt such as instruction framing, label semantics, or cultural grounding. Consequently, the role of prompt structure has become an under investigated area in multilingual reasoning.

\subsection{Prompting in Low-Resource, Cross-Lingual Situations.}
Prompting has been shown to rival parameter adaptation in certain low-resource scenarios. Few-shot cross-lingual studies demonstrate that direct in-language prompting or translate-then-prompt pipelines can match or outperform language-adaptive fine-tuning in several tasks \cite{toukmaji2023fewshot}. The language versioning in the prompting methods also indicates that the ability to induce capabilities can be raised without updating the model. \cite{nguyen2024linguistically}.

The work of the African-oriented models, including Lugha-LLaMA and InkubaLM, shows that the coverage of the language and timely may lead to performance gains (\cite{buzaaba2025lugha}, \cite{heath2024inkubalm}). Nevertheless, timely structure is not in the majority of instances regarded as an experimental variable; it is rather a fixed assessment framework.

\subsection{Cultural and Script-Aware Prompting}
African NLP datasets frequently involve culturally grounded semantics, orthographic variation, and code-mixing. Resources such as MasakhaNEWS \cite{adelani2023masakhanews} and AfriSenti \cite{muhammad2023afrisenti} demonstrate that in-language demonstrations can recover performance in classification tasks, but they do not systematically analyze native label semantics or culturally localized task framing.

In multilingual reasoning processes Orthography and script sensitivity are not well studied. Even though some studies indicate the use of orthographically explicit prompts to help with tasks like diacritics restoration \cite{ojo2025afrobench}, script and decision structure aware prompting has not been within African NLI.

\section{Prompting Strategies}
We evaluate five zero-shot prompting strategies designed to systematically vary linguistic grounding and decision structure while keeping the task formulation constant. All strategies require the model to output exactly one English label (entailment, contradiction, or neutral) without explanations. Full prompt templates are provided in the Appendix. Notably, all of the strategies do not depend on few-shot demonstrations and Chain-of-Thought (CoT) reasoning. We hope to the effect of the prompt alone.

\subsection{Baseline (Zero-Shot) Prompting}
The premise and hypothesis are presented directly through the baseline prompt and ask the model to decide on one of the three NLI labels. The description of the task is very minimalistic and impartial, presenting the available labels without a further linguistic or interpretative help. This formulation acts as the control scenario and all the structured prompting guidelines are compared against it.

\subsection{Language-Specific Prompting}
Language Specific prompting is an explicit form of reasoning, which puts the process in the cultural and pragmatic background of the target language. The model will be directed to read the sentences as an English native speaker, employing every day conception as opposed to systematic logical deduction. The prompt emphasizes:
\begin{itemize}
    \item Arguments like in ordinary speech,
    \item Taking into account what a common speaker himself would intuit,
    \item The use of general common sense and common sense practicality that is language-specific.
\end{itemize}
Decision rules are determined using terms of how a native speaker would accept, reject or be uncertain within by the premise. It is an exploratory method of whether culturally and linguistically situated grounded reasoning can promote semantic matching and stop over interpretation of unnatural logical reasoning.

\subsection{Contrastive Prompting}
Contrastive prompts organize the decision process to have all three of the possible interpretations clearly stated then the model chooses one. When comparing two items, the prompt does not request this directly; instead, it directs the model to make a comparison:
\begin{itemize}
    \item True or false of the premise, of the hypothesis.
    \item Whether it makes it false,
    \item Or insures it not or it opposes it.
\end{itemize}
This strategy will minimize premature label bias, and maximize balanced class selection by making the model non-deterministic, whenever it opts to give a response. The systematic comparison is assumed to promote more conscious discrimination within the three categories of NLI.

\subsection{Native-Label Self-Translation Prompting (NL-STP)}
Native-Label Self-Translation Prompting introduces a two-stage reasoning constraint. The model is instructed to:
\begin{itemize}
    \item Reason entirely in the target language,
    \item Take the best decision term in that language,
    \item Only the decision word selected is to be translated into English,
    \item Output exactly one English label.
\end{itemize}
This method aims at minimizing cross linguistic semantic mismatch between reasoning space and label space. By encouraging internal reasoning in the target language before mapping to English labels, NL-STP tests whether linguistic decoupling improves inference reliability in multilingual settings.

\subsection{Script-Aware Prompting}
Script aware prompting points the reasoning process squarely to the linguistic script element of the input. The model is told to take the text literally and then reason and make a choice on the label taken in the target language. Though our experiments are based on Latin script data as a main source, such a strategy is used to study whether the explicit reinforcement of script and context of language has an effect on the stability of inference.

Collectively, these strategies allow us to analyze how linguistic grounding, cultural framing, contrastive decision structuring, and cross-lingual label mapping influence multilingual NLI behavior in low-resource African languages.

\section{Experimental Setup}

\subsection{Dataset}
We evaluate our prompting strategies on the AfriXNLI benchmark, a multilingual Natural Language Inference (NLI) dataset covering several African languages. We are targeting three languages; Swahili, Yoruba, and Hausa. in all languages, the full test set of 600 examples (equally balanced by the three labels entailment, contradiction, and neutral 200 instances each) are used. Subsampling is not practiced as the size of the test set is moderate and balanced. All reported results are computed over the complete test split for each language.

\subsection{Models}
We evaluate two mid-sized open-weight large language models:
\begin{itemize}
    \item Llama3.2-3B
    \item Gemma3-4B
\end{itemize}
We are not interested in attaining state of the art performance, but rather examining prompting behavior under resource constrained environments. The models chosen are realistic deployment cases based on low resources settings where large scale models might not be available.

All the experiments are done in a zero-shot setup unless indicated otherwise. Neither fine-tuning, training of adapters or parameter updates are carried out.

\subsection{Prompting Protocol}
In Section 3, each of the prompting strategies is individually assessed within every language and model combination. The models are asked to provide only one English label entailment, contradiction or neutral as it is explained. Generation settings are kept deterministic (temperature = 0) to ensure reproducibility and reduce stochastic variability in label outputs. There are no few-shot demonstrations or Chain-of-Thought reasoning that are applied in the primary evaluation.

For comparative analysis, we additionally evaluate a baseline augmented with few-shot examples and Chain-of-Thought prompting to assess whether structured zero-shot prompts can outperform stronger reasoning-enhanced baselines.

\subsection{Evaluation Metrics}
We report:
\begin{itemize}
    \item Accuracy
    \item Macro-F1
    \item Per-class F1 scores (Entailment, Contradiction, Neutral)
\end{itemize}
Since there was a balanced distribution of the classes, Macro-F1 is a indicator of categories. Class F1 of the form that is especially significant in our study has a high prediction skew and a neutral class collapse in a number of the prompting strategies. We thus lay stress on class wise analysis as opposed to using only on aggregate accuracy.

\subsection{Analysis Design}
Our analysis focuses on three dimensions:
\begin{itemize}
    \item Predication distribution and class wise behavior changes.
    \item Cross-language consistency of prompting strategies
    \item Comparison between structured zero-shot prompting and stronger few-shot + CoT baselines
\end{itemize}
We study the influence of prompt formulation per se on the multilingual inference behavior of low-resource African languages under controlled model size and dataset to understand the role of prompt formulation alone in multilingual inference behavior.

\section{Results}
We evaluate five prompting strategies across three languages (Swahili, Yoruba, Hausa) and two mid sized open weight models (Llama3.2-3B and Gemma3-4B). Metrics are reported on complete 600 example test split of each language. The overall results are summarized in the Table 1 of Appendix, which reports accuracy and macro-F1 score values across all language model combinations. Detailed per class behavior and prediction distributions are also availabel in the Table 2 of Appendix.

\begin{figure}[h]
    \centering
    \includegraphics[width=1\linewidth]{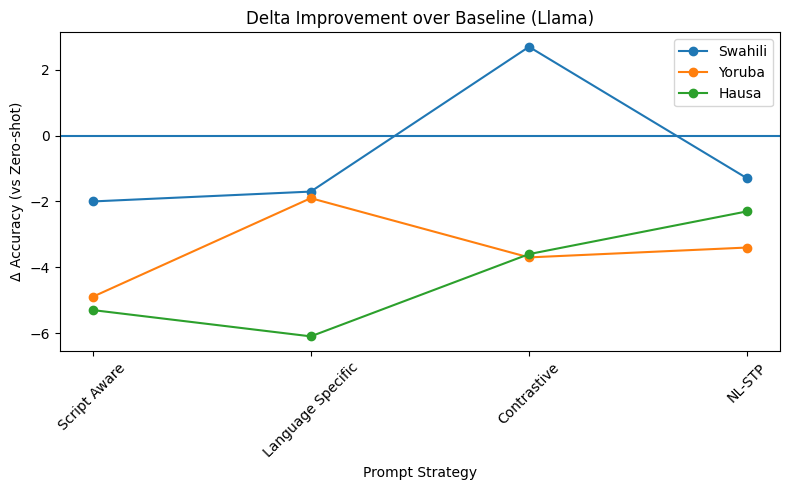}
    \caption{Delta accuracy improvement over the zero-shot baseline for Llama3.2-3B across prompting strategies and languages. Positive values indicate gains over baseline; negative values indicate degradation. The most consistent are improvements brought by contrastive prompting (across languages).}
    \label{fig:placeholder}
\end{figure}

\begin{figure}[h]
    \centering
    \includegraphics[width=1\linewidth]{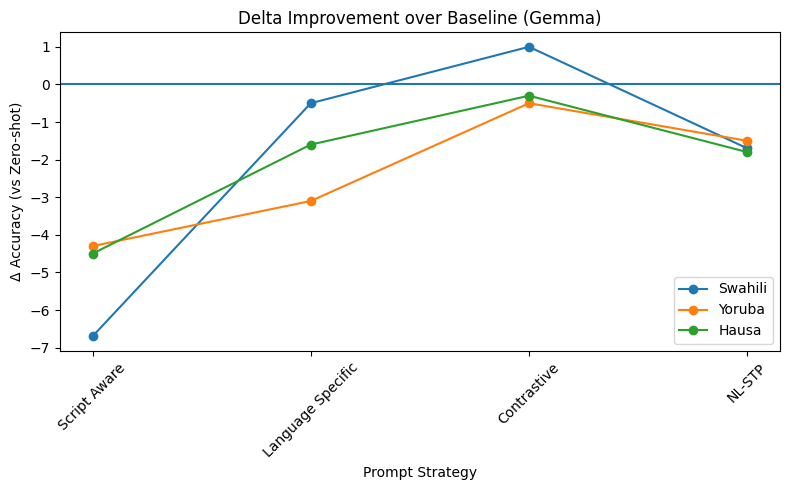}
    \caption{Table of delta accuracy improvement of Gemma3-4B between zero-shot base and prompting strategies in both languages. Contrastive prompting shows stable cross-language gains compared to other structured prompts.}
    \label{fig:placeholder}
\end{figure}

\section{Analysis}

\subsection{Overall Performance}
In all languages and models, there are significant variations between performance based on prompt formulation. The zero-shot prompt baseline has a moderate level of accuracy, but contains strong class imbalance tendencies in certain environments. Systematic strategies of prompting change both the overall performance as well as the performance of the classes. Figure 1 indicates that the size and direction of gains are much larger across strategies, although Contrastive prompting shows the most more stable performance as shown in Table 1. Figure 2 demonstrate a corresponding trend in a similar fashion as Gemma3-4B and the fact that contrastive decision framing is an inter architectural generalization. Among the considered methods, For Llama3.2-3B, Contrastive prompting raises the accuracy on Swahili by 34.5\% (baseline) to 37.2, and macro-F1 by 0.29 to 0.36 (Refer Table 1 of appendix). For Gemma3-4B, Contrastive prompting generates the highest accuracy in Swahili 42.2\% and second highest accuracy in Yoruba 38.3\%. In most environments contrastive prompting attains competitive performance but sometimes even baseline prompt performs better than it.

Language specific prompting and Script aware prompting present conflicting results. Although at times they do serve to improve performance on particular language model pairs, they also bring with them higher prediction skew on other models. The Native-Label Self-Translation Prompting (NL-STP) has shown better results in some settings, nevertheless, it shows instability and not much reliability.

\subsection{Class Wise Behavior}
Per class F1 comparison (Refer Table 2 of Appendix) indicates that models are biased towards prediction of a specific class like Neutral class here. The model is avoiding predicting Contradiction entirely in some cases that's why F1 score of Contradiction label is collapsing to 0 in some cases. In certain languages, even the zero-shot prompt at the baseline already has some imbalanced features. Prompts used systematically may either worsen or alleviate this behavior. Many cases showing huge counts in Neutral class prediction but very few for Contradiction label and patterns like 145/0/455, 51/0/549, etc. for Neutral/Contradiction/Entailment are observed (Refer Table 2 of Appendix).

Language specific prompting is showing more balanced F1 across classes whereas more stable performance across classes was observed in cases where Contrastive prompting was used. Zero-shot prompting on the other hand showed decent overall but highly skewed predictions. Prompt design here is not only improving accuracy but it's also changing the class behavior here.

\subsection{Cross Language and Cross Model Trends}
The patterns of performance vary in languages. The gains made in the Swahili language are not necessarily extended to the Yoruba and Hausa languages which means that the fast effectiveness depends on the linguistic features. On a model-to-model comparison, Gemma3-4B is considered to have a fraction more stable classwise behavior than Llama3.2-3B, yet they both are sensitive to prompt structure. It is also important to note that Contrastive prompting gives rather fixed gains in both architectures indicating an absence of strong model dependence.

\begin{figure}[h]
    \centering
    \includegraphics[width=1\linewidth]{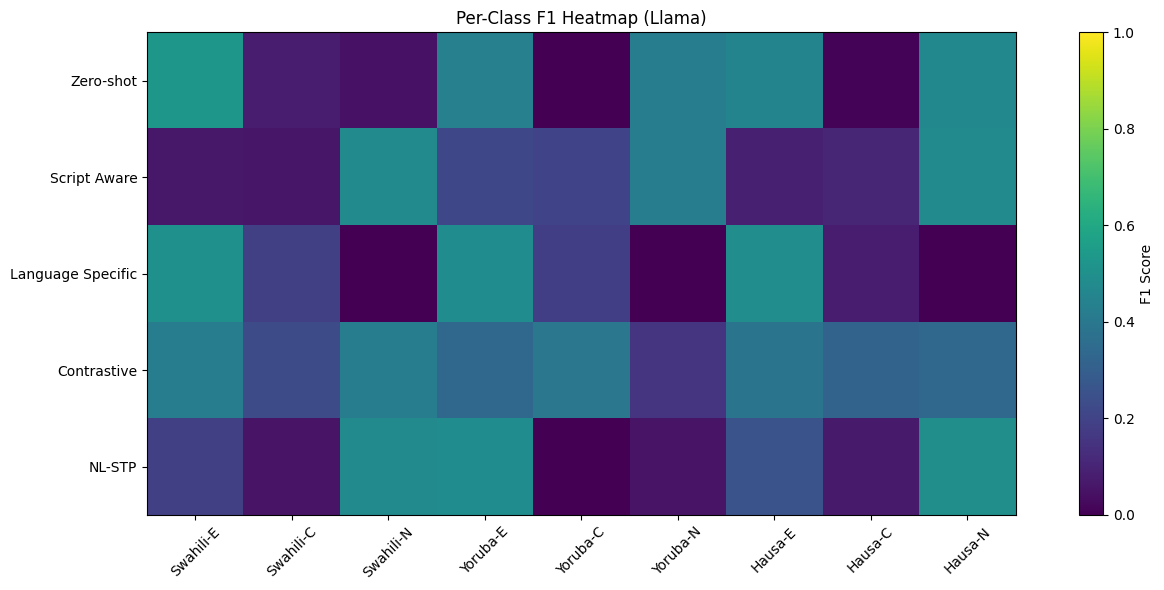}
    \caption{Per-class F1 heatmap of Gemma3-4B, between prompting strategies, languages and labels (E, C, N). The higher the values are darker, the better the performance on the basis of classes. Under a variety of prompting set-ups, neutral-class instability is apparent.}
    \label{fig:placeholder}
\end{figure}

\begin{figure}[h]
    \centering
    \includegraphics[width=1\linewidth]{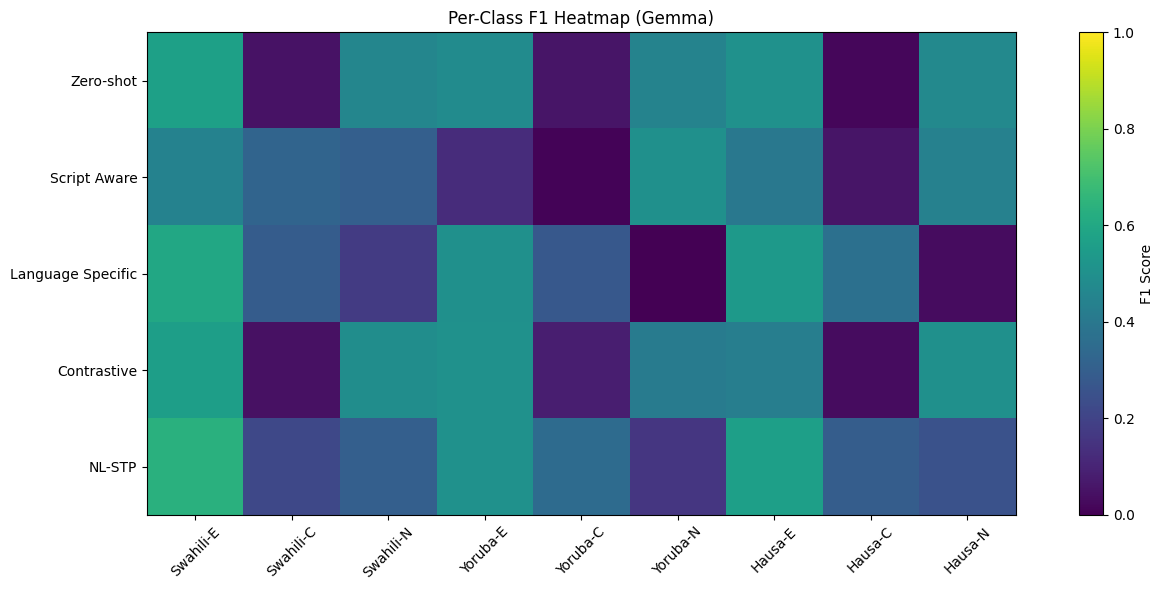}
    \caption{Per-class F1 heatmap for Llama3.2-3B across prompting strategies, languages, and labels. Compared to Gemma3-4B, Llama exhibits stronger sensitivity to prompt formulation and greater class imbalance in certain configurations.}
    \label{fig:placeholder}
\end{figure}

\subsection{Best Strategy vs Baseline (Zero-Shot)}
The behavior of the classes in the strategies is graphed in Figure 3 in Gemma3-4B. The same result is illustrated in Figure 4, which shows even more instability on the case of Llama3.2-3B. In any language and model, Contrastive prompting appears to be the most stable and progressively improving strategy compared to the base. Despite language specific variation in the magnitude of gained absolute accuracy, Contrastive prompting decreases extreme prediction skew in most settings, and increases macro-F1. The imbalance behavior is common to the baseline prompt it tends to overestimate the neutral label or simply collapses into a dominant label. Conversely, Structured comparison proposed by Contrastive prompting facilitates a more equal assessment between entailment, contradiction and neutral. This results in better class-wise F1 stability, where the gains in overall accuracy may seem small.

Meanwhile, it is important to note that not all languages are equally improved. Certain model configurations have stronger gains in Swahili and Hausa than in Yoruba, indicating that the interaction between prompt model effectiveness and linguistic structure and model representations is evident. Contrastive prompting however does not bring any drastic regressions in any case which supports its strong presence.

\subsection{Neutral Class Collapse and Prediction Skew}
Neutral class collapse in which the ground truth is balanced, but the models all predict neutral cases badly, is a common recurring theme of numerous prompting strategies. Language specific prompting with Llama3.2-3B Yoruba and Hausa showed neutral F1 equal to 0.00 although the ground-truth is balanced (200 instances of neutral). Conversely, Contrastive prompting raises neutral F1 to 0.42 (Swahili), 0.15 (Yoruba) and 0.33 (Hausa). In the case of Gemma3-4B, the base neutral F1 of Yoruba is 0.45, but the value implodes to 0.00 when prompted to by Language Specific prompting but rebounds to 0.41 after contrastive prompting. This tendency is particularly noticeable in the baseline and some language specific settings, in which per-class F1 of the neutral approaches acquires zero. This is well depicted in Figure 5.

\begin{figure}[h]
    \centering
    \includegraphics[width=1\linewidth]{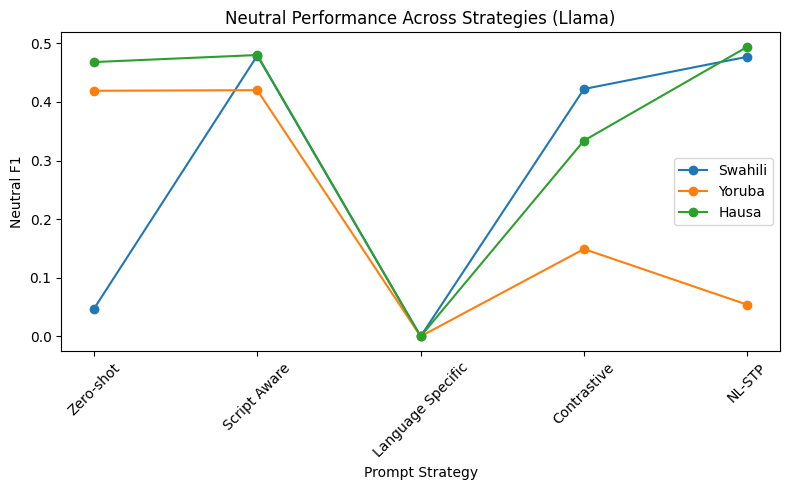}
    \caption{F1 of neutral between prompting strategies of Llama3.2-3B. There are a number of strategies that are of neutral-class collapsing, and Contrastive prompting has more stable neutral performance.}
    \label{fig:placeholder}
\end{figure}

Figure 6 shows a similar but slightly more stable pattern for Gemma3-4B.

\begin{figure}[h]
    \centering
    \includegraphics[width=1\linewidth]{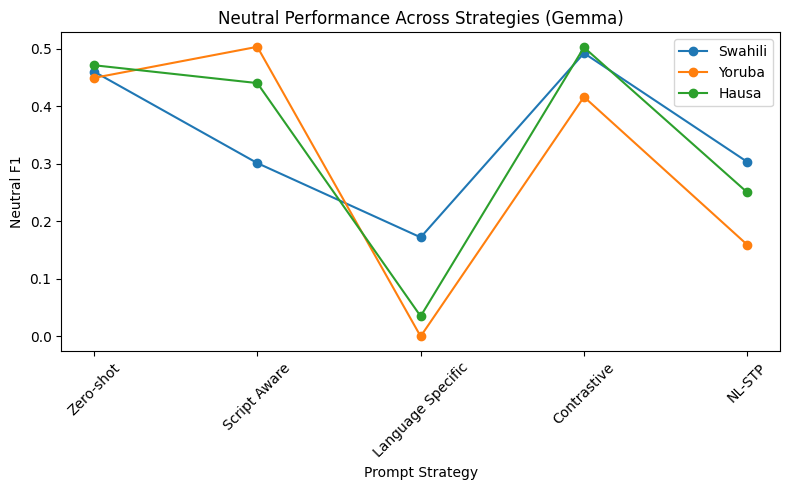}
    \caption{F1 of neutral class with respect to prompting strategies of Gemma3-4B. The occurrence of neutral collapse is reduced by contrastive prompting as compared to other strategies.}
    \label{fig:placeholder}
\end{figure}

The method of Language specific prompting though this rests on pragmatic reasoning, there are actually occasions when language specific prompting enhances the predictions of entailment at the cost of neutral stability. The NL-STP in some cases will increase entailment/contradiction alignment, but it is not necessarily able to avoid collapse.

Contrastive prompting mitigates this phenomenon by explicitly framing the decision as a three way comparison. It seems to discourage premature commitment to a prevailing class by making the model consider competitive interpretations first of all. This structured discrimination leads to more balanced prediction distributions across languages.

\subsection{Cross-Language Consistency}
The instantaneous efficacy is diverse in Swahili, Yoruba, and Hausa. Some of the strategies used perform better in a particular language, but show insignificant or unstable results on others. This implies that it cannot be supposed that multilingual prompting can be uniformly transferred among related languages.

Nevertheless, where there were these differences, Contrastive prompting shows uniformity in every three languages. Cases of stability in its performance suggest that the structured decision framing can have better generalization than either culturally based or internally translated reasoning restrictions.

\subsection{Model level Observations}
Comparing architectures, Gemma3-4B exhibits slightly more stable class-wise behavior than Llama3.2-3B, particularly under structured prompting. Nevertheless, the two models are not insensitive to timeliness formulation. Relative stability of the Contrastive prompting in both models suggests that its effectiveness is in the structuring of decision making but not the peculiarities of the model.

\section{Discussion}
The results of our study indicate that the level of multilingual NLI in African languages under low resources is extremely delicate as to prompt formulation. Even where the general differences in accuracy are intermediate, there is significant restructuring of the behavior and prediction distributions of classes by prompt structure. This implies that assessment of multilingual environment needs to be performed to look past aggregate accuracy and scrutinize calibration and label stability to a greater degree.

One of the main findings in our work is that as a result of a number of prompting arrangements, the tendency to occur with a neutral-class collapse is predominant. Even with equal distributions of classes in AfriXNLI, models do not predict neutral often, and when it is based on minimal or pragmatically formulated prompts. This observation shows that in case of underspecified task instructions, there is a tendencies of unpredictable labels that default toward more decisive interpretations (entailment or contradiction). Such bias is even stronger in low-resource languages, where model pretraining data and target text may already contain semantic mist match.

Contrastive prompting consistently mitigates this instability. By explicitly presenting competing interpretations before requiring a decision, it introduces a lightweight structural constraint that encourages more balanced reasoning. This suggests that decision framing not additional demonstrations or longer reasoning chains can play a critical role in improving inference robustness. In resource constrained environments, where few-shot curation and Chain-of-Thought prompting may be impractical or computationally expensive, structured zero-shot prompting offers a compelling alternative.

In a broader sense, our results will answer the research questions set out in the Section 1:
\begin{itemize}
    \item {RQ1:} Prompt design has a strong impact on the inference behavior in the classes, regularly reforming the prediction distributions even in cases where the difference in accuracy would be moderate.
    \item {RQ2:} Contrastive prompting mitigates prediction skew and reduces neutral-class collapse, demonstrating improved calibration across labels.
    \item {RQ3:} The carefully designed prompts may compete or be better at a few shots of CoT in multilingual resource-constrained settings of NLI.
\end{itemize}

In general, the discussion shows that the performance of multilingual NLI in low-resource languages of Africa is very sensitive related to the timely creation. Comparison-based prompting performed in a structured way provides the most stable and robust behavior implying that decision framing is an important factor in multilingual inference reliability.

These findings highlight that timely engineering in a multilingual environment must not only aim at achieving superior accuracy, but also strive to accommodate equal, steady and linguistically based inference action.

\section{Conclusion}
In AfriXNLI benchmark and with two open-weight models with middle size, we performed a systematic survey of zero-shot prompting methods on Natural Language Inference in Swahili, Yoruba, and Hausa. Our results show that prompt design significantly shapes class-wise behavior and prediction stability, even when overall accuracy differences are modest.

We observe frequent neutral-class collapse and prediction skew under several prompting formulations, highlighting the importance of analyzing per-class performance rather than relying solely on aggregate metrics. Among the evaluated strategies, Contrastive prompting emerges as the most stable and consistently more stable across languages and models.

Importantly, carefully structured zero-shot prompts can match or outperform stronger baselines augmented with few-shot examples and Chain-of-Thought reasoning. In general, our results indicate that in low-resource African languages, such things as decision framing and designing language-adjusted prompts are of crucial importance in the reliability of multilingual NLI. Future work should feature more curated evaluation sets; more African languages are constantly undergoing efforts in terms of expanding their dataset to NLI-style datasets along with exploring interactions between structured prompting and lightweight adaptation methods.

\section{Ethical Considerations and Limitations}
\begin{itemize}
    \item Low coverage of the languages: We only assess three African languages (Swahili, Yoruba, Hausa). The finding might not be generalizable to other low-resource languages.
    \item Single benchmark: Experiments are done on the AfriXNLI only. It would be improved by testing other NLI data.
    \item Model scale constraints: We use mid-sized open-weight models (3–4B). Larger or multilingual foundation models may exhibit different prompting sensitivities.
    \item No qualitative error analysis: We emphasize on quantitative measures. A more detailed analysis of the error on an instance-level might be used as a way of gaining a better understanding of the phenomenon of neutral-class collapse and semantic misalignment.
    
\end{itemize}
\section{Bibliographical References}
\bibliographystyle{lrec2026-natbib}
\bibliography{lrec2026-example}

\appendix
\section{Appendix - Prompt Templates}

This section presents the exact prompt templates used for each strategy in our experiments where \texttt{\{lang\}}, \texttt{\{premise\}}, and \texttt{\{hypothesis\}} are replaced at runtime.

\subsection{Language-Specific Prompting}

{\small
\begin{verbatim}
PROMPT = """
You are interpreting these sentences 
as a native {lang} speaker, using 
everyday {lang} understanding and 
cultural context.

INSTRUCTIONS:
1) Reason as a native {lang} speaker 
would in daily conversation, not 
using formal logic.
2) Consider what a typical speaker 
would naturally infer from the first 
sentence about the second.
3) Decide the relationship based on 
common-sense and pragmatic 
understanding in {lang}.
4) Output exactly ONE English word: 
entailment, contradiction, or 
neutral.
5) Do NOT output explanations or 
any extra text.

Decision rules (according to 
native {lang} usage):
- entailment: a typical {lang} 
speaker would accept the second 
sentence as true because of the 
first.
- contradiction: a typical {lang} 
speaker would judge the second 
sentence as incompatible with the 
first.
- neutral: a typical {lang} speaker 
would find that the first does not 
clearly determine the second.

Premise: "{premise}"
Hypothesis: "{hypothesis}"
Answer:
"""
\end{verbatim}
}

\subsection{Contrastive Prompting}

{\small
\begin{verbatim}
PROMPT = """
You are comparing three possible 
interpretations of the relationship 
between the following sentences.

INSTRUCTIONS:
1) Consider each of the three 
possibilities below.
2) Decide which one best matches 
the relationship between the 
sentences.
3) Output exactly ONE English 
word: entailment, contradiction, 
or neutral.
4) Do NOT output explanations or 
any extra text.

Interpretations:
- entailment: the premise makes 
the hypothesis true.
- contradiction: the premise 
makes the hypothesis false.
- neutral: the premise neither 
guarantees nor contradicts the 
hypothesis.

Premise: "{premise}"
Hypothesis: "{hypothesis}"

Which interpretation fits best?
Answer:
"""
\end{verbatim}
}

\subsection{Native-Label Self-Translation Prompting (NL-STP)}

{\small
\begin{verbatim}
PROMPT = """
You must decide the relationship 
using the target language first, 
and only then map it to English.

INSTRUCTIONS:
Step 1: Read the sentences and 
reason entirely in {lang}.
Step 2: Choose the most 
appropriate decision word in 
{lang}.
Step 3: Translate ONLY that 
chosen decision word into 
English.
Step 4: Output exactly ONE 
English word: entailment, 
contradiction, or neutral.

Do NOT output explanations 
or any other text.

Premise: "{premise}"
Hypothesis: "{hypothesis}"
Final Answer (English, one 
word only):
"""
\end{verbatim}
}

\subsection{Baseline (Zero-Shot)}

{\small
\begin{verbatim}
PROMPT = """
Given the premise and 
hypothesis, determine their 
relationship.

Choose exactly one of the 
following:
- entailment
- contradiction
- neutral

Premise: "{premise}"
Hypothesis: "{hypothesis}"
Answer:
"""
\end{verbatim}
}

\subsection{Script aware Prompting}

\textbf{Ajami Variant:}

{\small
\begin{verbatim}
PROMPT = """
The following text is written 
in the Arabic-derived Ajami 
script used for {lang}.

INSTRUCTIONS:
1) Internally transliterate 
the text into {lang} written 
in Latin script. Do NOT output 
it.
2) Reason in {lang}.
3) Decide the relationship.
4) Output exactly ONE English 
word: entailment, contradiction, 
or neutral.

Decision rules:
- entailment: premise makes 
hypothesis true.
- contradiction: premise makes 
hypothesis false.
- neutral: neither true nor 
false.

Premise (Ajami): "{premise}"
Hypothesis (Ajami): "{hypothesis}"
Answer:
"""
\end{verbatim}
}

\textbf{Latin Script Variant:}

{\small
\begin{verbatim}
PROMPT = """
You are a fluent {lang} speaker.

INSTRUCTIONS:
1) Read and reason in {lang}.
2) Decide the relationship.
3) Output exactly ONE English 
word: entailment, contradiction, 
or neutral.

Decision rules:
- entailment: premise makes 
hypothesis true.
- contradiction: premise makes 
hypothesis false.
- neutral: neither true nor 
false.

Premise: "{premise}"
Hypothesis: "{hypothesis}"
Answer:
"""
\end{verbatim}
}

\section{Appendix - Full Results}
\begin{table*}[t]
\centering
\small
\renewcommand{\arraystretch}{0.6}

\begin{adjustbox}{width=\textwidth}
\begin{tabular}{l l l c c}
\toprule
Strategy & Lang & Model & Acc & Macro-F1 \\
\midrule

Script Aware & Sw & Llama & 0.325 & 0.202 \\
Script Aware & Sw & Gemma & 0.345 & 0.356 \\
Script Aware & Yo & Llama & 0.308 & 0.279 \\
Script Aware & Yo & Gemma & 0.345 & 0.213 \\
Script Aware & Ha & Llama & 0.330 & 0.225 \\
Script Aware & Ha & Gemma & 0.353 & 0.300 \\

Language Spec. & Sw & Llama & 0.328 & 0.231 \\
Language Spec. & Sw & Gemma & 0.407 & 0.355 \\
Language Spec. & Yo & Llama & 0.338 & 0.224 \\
Language Spec. & Yo & Gemma & 0.357 & 0.258 \\
Language Spec. & Ha & Llama & 0.322 & 0.189 \\
Language Spec. & Ha & Gemma & 0.382 & 0.314 \\

Contrastive & Sw & Llama & 0.372 & 0.357 \\
Contrastive & Sw & Gemma & 0.422 & 0.365 \\
Contrastive & Yo & Llama & 0.320 & 0.294 \\
Contrastive & Yo & Gemma & 0.383 & 0.335 \\
Contrastive & Ha & Llama & 0.347 & 0.345 \\
Contrastive & Ha & Gemma & 0.395 & 0.322 \\

NL-STP & Sw & Llama & 0.332 & 0.241 \\
NL-STP & Sw & Gemma & 0.395 & 0.385 \\
NL-STP & Yo & Llama & 0.323 & 0.179 \\
NL-STP & Yo & Gemma & 0.373 & 0.338 \\
NL-STP & Ha & Llama & 0.360 & 0.274 \\
NL-STP & Ha & Gemma & 0.380 & 0.371 \\

Zero-shot & Sw & Llama & 0.345 & 0.218 \\
Zero-shot & Sw & Gemma & 0.412 & 0.360 \\
Zero-shot & Yo & Llama & 0.357 & 0.285 \\
Zero-shot & Yo & Gemma & 0.388 & 0.329 \\
Zero-shot & Ha & Llama & 0.383 & 0.309 \\
Zero-shot & Ha & Gemma & 0.398 & 0.332 \\

\bottomrule
\end{tabular}
\end{adjustbox}

\caption{Accuracy and Macro-F1 across prompting strategies, languages, and models.}
\end{table*}

\begin{table*}[t]
\centering
\small
\begin{adjustbox}{width=\textwidth}
\begin{tabular}{llllll}
\toprule
Strategy & Lang & Model & F1 (E/C/N) & Pred (C/N/E) \\
\midrule

Script Aware & Sw & Llama & 0.064/0.062/0.479 & 26/21/18 \\
Script Aware & Sw & Gemma & 0.444/0.324/0.301 & 226/218/102 \\
Script Aware & Yo & Llama & 0.217/0.200/0.420 & 130/134/104 \\
Script Aware & Yo & Gemma & 0.126/0.010/0.503 & 2/6/39 \\
Script Aware & Ha & Llama & 0.087/0.108/0.480 & 41/216/30 \\
Script Aware & Ha & Gemma & 0.401/0.058/0.440 & 40/28/114 \\

Language Spec. & Sw & Llama & 0.501/0.191/0.000 & 145/0/455 \\
Language Spec. & Sw & Gemma & 0.601/0.292/0.172 & 183/68/349 \\
Language Spec. & Yo & Llama & 0.486/0.187/0.000 & 68/0/532 \\
Language Spec. & Yo & Gemma & 0.501/0.274/0.000 & 114/4/482 \\
Language Spec. & Ha & Llama & 0.489/0.080/0.000 & 51/0/549 \\
Language Spec. & Ha & Gemma & 0.537/0.370/0.035 & 238/26/336 \\

Contrastive & Sw & Llama & 0.418/0.230/0.422 & 105/260/235 \\
Contrastive & Sw & Gemma & 0.559/0.043/0.492 & 36/414/137 \\
Contrastive & Yo & Llama & 0.338/0.396/0.149 & 305/69/226 \\
Contrastive & Yo & Gemma & 0.505/0.085/0.416 & 35/329/236 \\
Contrastive & Ha & Llama & 0.383/0.317/0.334 & 166/195/239 \\
Contrastive & Ha & Gemma & 0.427/0.035/0.502 & 26/465/109 \\

NL-STP & Sw & Llama & 0.191/0.054/0.477 & 22/490/82 \\
NL-STP & Sw & Gemma & 0.634/0.217/0.303 & 168/203/229 \\
NL-STP & Yo & Llama & 0.485/0.000/0.054 & 0/19/576 \\
NL-STP & Yo & Gemma & 0.506/0.351/0.159 & 205/77/318 \\
NL-STP & Ha & Llama & 0.255/0.074/0.494 & 17/492/90 \\
NL-STP & Ha & Gemma & 0.566/0.295/0.250 & 186/176/238 \\

Zero-shot & Sw & Llama & 0.529/0.077/0.047 & 61/0/526 \\
Zero-shot & Sw & Gemma & 0.570/0.049/0.460 & 46/361/193 \\
Zero-shot & Yo & Llama & 0.437/0.000/0.419 & 0/0/299 \\
Zero-shot & Yo & Gemma & 0.482/0.055/0.449 & 19/374/207 \\
Zero-shot & Ha & Llama & 0.450/0.010/0.468 & 4/1/240 \\
Zero-shot & Ha & Gemma & 0.505/0.018/0.471 & 21/407/172 \\

\bottomrule
\end{tabular}
\end{adjustbox}
\caption{Per-class F1 scores and prediction distributions.}
\end{table*}

\end{document}